\RequirePackage{fix-cm}
\documentclass[smallextended, final]{svjour3}       
\journalname{Data Mining and Knowledge Discovery}
\smartqed  
\usepackage{graphicx}
\usepackage{natbib}
\usepackage{comment}
\usepackage{url}

\usepackage{amsmath}
\usepackage{algorithm}
\usepackage{algpseudocode}
\usepackage{array}
\begin{document}

\title{MrSQM: Fast Time Series Classification with Symbolic Representations and Efficient Sequence Mining 
}

\titlerunning{MrSQM: Fast Time Series Classification with Symbolic Representations}        

\author{Thach Le Nguyen \and 
        Georgiana Ifrim
}


\institute{Thach Le Nguyen and Georgiana Ifrim are with the School of Computer Science, Insight Centre for Data Analytics, 
        University College Dublin, Ireland. \\
        \{thach.lenguyen, georgiana.ifrim\}@ucd.ie
}
\date{Received: date / Accepted: date}

\maketitle

\begin{abstract}


Symbolic representations of time series have proven to be effective for time series classification, with many recent approaches including BOSS, WEASEL, and MrSEQL. The key idea is to transform numerical time series to symbolic representations in the time or frequency domain, i.e., sequences of symbols, and then extract features from these sequences. While achieving high accuracy, existing symbolic classifiers are computationally expensive. It is also not clear whether further accuracy and speed improvements could be gained by a careful analysis of the symbolic transform and the trade-offs between time domain and frequency domain symbolic features.
In this paper we present MrSQM, a new time series classifier that uses  multiple symbolic representations and efficient sequence mining, to extract important time series features. We study two symbolic transforms and four feature selection approaches on symbolic sequences, ranging from fully supervised, to unsupervised and hybrids. We propose a new approach for optimal supervised symbolic feature selection in all-subsequence space, by adapting a Chi-squared bound developed for discriminative pattern mining, to time series. Our experiments on the 112 datasets of the UEA/UCR benchmark demonstrate that MrSQM can quickly extract useful features and learn accurate classifiers with the  logistic regression algorithm. We show that a fast symbolic transform combined with a simple feature selection strategy can be highly effective as compared to more sophisticated and expensive feature selection methods. MrSQM completes training and prediction on 112 UEA/UCR datasets in 1.5h for an accuracy comparable to existing efficient state-of-the-art methods, e.g., MrSEQL (10h) and ROCKET (2.5h). Furthermore, MrSQM enables the user to trade-off accuracy and speed by controlling the type and number of symbolic representations, thus further reducing the total runtime to 20 minutes for a similar level of accuracy.

\end{abstract}
\section{Introduction}
\label{sec:intro}

Symbolic representations of time series are a family of techniques to transform numerical time series to sequences of symbols, and were shown to be more robust to noise and useful for building effective time series classifiers.
Two of the most prominent symbolic representations are Symbolic Aggregate Approximation (SAX) \citep{lin-sax:dmkd07} and Symbolic Fourier Approximation (SFA) \citep{Schafer:2012:SSF:2247596.2247656}.
SAX-based classifiers include BOP \citep{lin-sax:dmkd07,Lin2012:iis}, FastShapelets \citep{fast-shapelet:sdm13}, SAX-VSM \citep{senin-saxvsm:icdm13}, while SFA-based classifiers include BOSS \citep{schafer2015boss}, BOSS VS \citep{schaefer:dmkd16} and WEASEL \citep{Schafer:2017:weasel}. MrSEQL \citep{LeNguyen2019} is a symbolic classifier which utilizes both SAX and SFA transformations, which further improved the accuracy and speed of classification. Several state-of-the-art ensemble methods, e.g., HIVE-COTE \citep{bagnall2016great,hive-cote,Bagnall2020ATO} and TS-CHIEF \citep{Shifaz2020TSCHIEFAS}, incorporate symbolic representations for  their constituent classifiers and are the current state-of-the-art with regard to accuracy.

Symbolic representations of time series enable the adoption of techniques developed for text mining. For example, SAX-VSM, BOSS VS and WEASEL make use of tf-idf vectors and vector space models \citep{senin-saxvsm:icdm13,schaefer:dmkd16}, while MrSEQL is based on a sequence learning algorithm developed for text classification  \citep{ifrim-seql:kdd11}. These apparently different approaches can be summarized as methods of extracting discriminative features from symbolic representations of time series, coupled with a classifier. 
While achieving high accuracy, the key challenge for symbolic classifiers is to efficiently select good features from a large feature space. For example, even with fixed parameters, a SAX bag-of-words can contain as many as $\alpha^w$ unique words, in which $\alpha$ is the size of the alphabet (the number of distinct symbols) and $w$ is the length of the words. Even for moderate alphabet and word sizes, this feature space grows quickly, e.g., for typical SAX parameters $\alpha=4, w=16$, there can be 4 billion unique SAX words.
SAX-VSM works with a single SAX representation, but the process for optimizing the SAX parameters is expensive. WEASEL has high accuracy by using SFA unigrams and bigrams but a high memory demand, due to needing to store all the SFA words before applying feature selection.
MrSEQL uses the feature space of all subsequences in the training data, in order to find useful features inside SAX or SFA words. It employs greedy feature selection and a gradient bound to quickly prune non-promising features. Despite these computational challenges, these methods are still vastly faster and less resource demanding than most state-of-the-art classifiers, in particular ensembles, e.g., HIVE-COTE, TS-CHIEF, and deep learning models, e.g., InceptionTime \citep{DBLP:journals/datamine/FawazLFPSWWIMP20}.

The recent time series classifiers ROCKET  \citep{dempster2019rocket} and MiniROCKET \citep{DBLP:conf/kdd/DempsterSW21} have again highlighted the effectiveness of methods which rely on large feature spaces and efficient linear classifiers. 
Like ROCKET, the approaches WEASEL and MrSEQL combine large feature spaces with linear classifiers, even though both employ different strategies to filter out the  features, and they work with symbolic representations rather than the raw time series data. These observations have inspired us to re-examine fast symbolic transforms, feature selection and linear classifiers for working with symbolic representations of time series. In particular, we are motivated by the effectiveness of using features based on the Fast Fourier Transform (e.g., as showcased in WEASEL and MrSEQL with SFA features) and the benefit we can gain from building on more than 20 years of research and implementations for improving the scalability and effectiveness of discrete Fourier transforms\footnote{Fastest Fourier Transform in the West: \url{https://www.fftw.org}}.

Our main contributions in this paper are as follows:
\begin{itemize}
    \item We propose \textbf{Multiple Representations Sequence Miner (MrSQM)}, a new symbolic time series classifier which builds on multiple symbolic representations, efficient sequence mining and a linear classifier, to achieve high accuracy with reduced computational cost.
    \item We study different feature selection strategies for symbolic representations of time series, including supervised, unsupervised and hybrids, and show that a very simple feature selection strategy is highly effective as compared with more sophisticated and expensive methods. 
    \item 
    We propose a new approach for supervised symbolic feature selection in all-subsequence space, by adapting a Chi-square bound developed for discriminative pattern mining, to time series. The bound guarantees finding the optimal features under the Chi-square test and enables us to find good features quickly.
    \item We present an extensive empirical study comparing the accuracy and runtime of MrSQM to recent state-of-the-art time series classifiers on 112 datasets of the new UEA/UCR TSC benchmark~\citep{bagnall2018uea}. 
    \item All our code and data is publicly available to enable reproducibility of our results\footnote{\url{https://github.com/mlgig/mrsqm}}. Our code is implemented in C++, but we also provide Python wrappers and a Python Jupyter Notebook with detailed examples to make the implementation more widely accessible.
    
 
\end{itemize}
 
The rest of the paper is organised as follows. In Section \ref{sec:relwork} we discuss the state-of-the-art in time series classification research. In Section \ref{section:method} we describe our research methodology. In Section \ref{sec:eval} we present an empirical study with a detailed sensitivity analysis for our methods and a comparison to state-of-the-art  time series classifiers. 
We conclude in Section \ref{sec:conc} with a discussion of strengths and weaknesses for our proposed methods.

\section{Related Work}
\label{sec:relwork}

The state-of-the-art in time series classification (TSC) has evolved rapidly with many different approaches contributing to improvements in  accuracy and speed.
The main baseline for TSC is 1NN-DTW \citep{bagnall2016great}, a one Nearest-Neighbor classifier with Dynamic Time Warping as distance measure. While this baseline is at times preferred for its simplicity, it is not very robust to noise and has been significantly outperformed in accuracy by more recent methods. Some of the most successful TSC approaches typically fall into the following three groups. 

\textbf{Ensemble Classifiers} aggregate the predictions of many independent classifiers. Each classifier is trained with different data representations and feature spaces, and the individual predictions are weighted based on the quality of the classifier on validation data. HIVE-COTE \citep{hive-cote} is the most popular example of such an approach. It is an evolution of the COTE \citep{bagnall2016great} ensemble and it is still currently the most accurate TSC approach.
While being very accurate, this method's runtime is bound to the slowest of its component classifiers. Recent work \citep{Bagnall2020ATO,DBLP:journals/ml/MiddlehurstLFLB21} has proposed techniques to make this approach more usable by improving its runtime, but it still requires more than two weeks to train on the new UEA/UCR benchmark which has a moderate size of about 300Mb. 
TS-CHIEF \citep{Shifaz2020TSCHIEFAS} is another recent ensemble which only uses decision tree classifiers. It was proposed as a more scalable alternative to HIVE-COTE, but it still takes weeks to train on the UEA/UCR benchmark. This makes the reproducibility of results with these methods challenging.

\textbf{Deep Learning Classifiers} were recently proposed for time series data and analysed in an extensive empirical survey \citep{IsmailFawaz2018deep}. Methods such as Fully Convolutional Networks (FCN) and Residual Networks (Resnet) were found to be highly effective and achieve accuracy comparable to HIVE-COTE. One issue with such approaches is their tendency to overfit with small training data and to have a high variance in accuracy. In order to create a more stable approach InceptionTime \citep{DBLP:journals/datamine/FawazLFPSWWIMP20} recently proposed to ensemble five deep learning classifiers. InceptionTime achieves an accuracy comparable to HIVE-COTE, but it requires vast computational resources and also takes days to train \citep{DBLP:journals/datamine/FawazLFPSWWIMP20,Bagnall2020ATO}.

\textbf{Linear Classifiers} 
were recently shown to work well for time series classification. Given a large feature space, the need for further feature expansion and learning non-linear classifiers is reduced. This idea was incorporated very successfully for large scale classification in libraries such as LIBLINEAR \citep{liblinear08}. In the context of TSC, this idea was incorporated by classifiers such as WEASEL \citep{Schafer:2017:weasel}, which creates a large SFA-words feature space, filters it with Chi-square feature selection, then learns a logistic regression classifier. Another linear classifier, MrSEQL \citep{LeNguyen2019}, uses a large feature space of SAX and SFA subwords, which is filtered using greedy gradient descent and logistic regression. A recent classifier ROCKET \citep{dempster2019rocket} generates many random convolutional kernels 
and uses max pooling and a new feature called \textit{ppv} to capture good features from the time series. ROCKET uses a large  feature space of 20,000 features (default settings) associated with the kernels, and a linear classifier (logistic regression or ridge regression). MiniROCKET \citep{DBLP:conf/kdd/DempsterSW21} is a very recent extension of ROCKET with comparable accuracy and faster runtime. 
These approaches were shown to be as accurate as ensembles and deep learning for TSC, but are orders of magnitude faster to train \citep{LeNguyen2019,dempster2019rocket,DBLP:journals/ml/MiddlehurstLFLB21}. MrSEQL can train on the UEA/UCR benchmark in 10h, while ROCKET has further reduced this time to under 2h.
Another advantage of these methods is their conceptual simplicity, since the method can be broken down into three stages: (1) transformation (e.g., symbolic for WEASEL and MrSEQL, or convolutional kernels for ROCKET), (2) feature selection and (3) linear classifier. 
Intuitively, these methods extract many shapelet-like features from the training data, and use the linear classifier to learn weights to filter out the useful features from the rest.
While there is a vast literature on shapelet-learning techniques, e.g., \citep{ye-shapelets:kdd09,fast-shapelet:sdm13,grabocka-lts:kdd14,bagnall2016great,Bagnall2020ATO} these recent linear classification methods were shown to be more accurate and faster than other shapelet-based approaches. In particular, the SFA transform does not require data normalisation (which may harm accuracy for some problems), it was shown to be robust to noise \citep{schafer2015boss}, and has very fast implementations that can further benefit from the past 20 years of work on speeding up the computation of Discrete Fourier Transform \citep{FFTW05,fftw21}.

Based on these observations and the recent success of symbolic transforms and linear classifiers, we focus our work on designing and evaluating new TSC methods built on   large symbolic feature spaces and efficient linear classifiers.
\section{Proposed Method}
\label{section:method}

\begin{center}
\begin{figure}[ht]
\includegraphics[width=\linewidth]{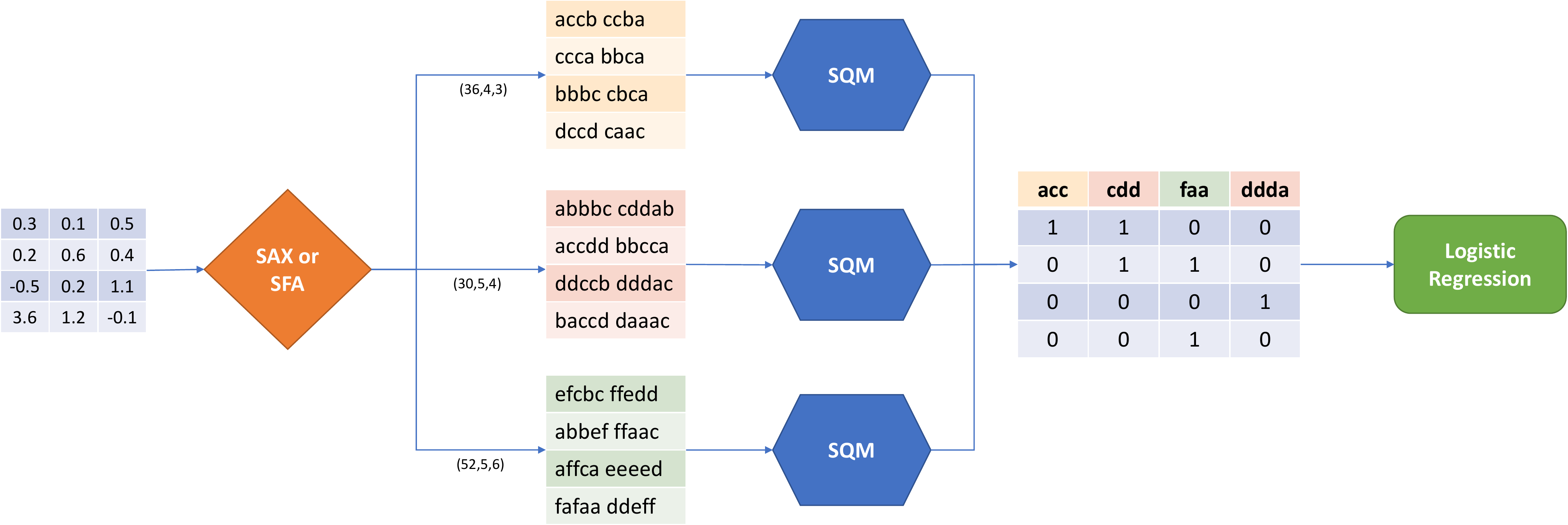}
\caption{Workflow for the MrSQM time series classifier with 3 stages: 1. symbolic transform, 2. feature selection, 3. classifier learning.}
\label{fig:mrsqm} 
\end{figure}
\end{center}

The MrSQM time series classifier has three main building blocks: (1) \textbf{symbolic transformation}, (2) \textbf{feature selection} and (3) \textbf{learning algorithm} for training a classifier. In the first stage, we transform the numerical time series to multiple symbolic representations using either SAX or SFA transforms. We carefully analyse the impact of parameter selection for the symbolic transform, as well as integrate fast transform implementations, especially for the discrete Fourier transform in SFA. 
For the second stage, to extract subsequence features from the symbolic representations, we explore new ideas for efficient feature selection with both supervised and unsupervised approaches. In particular, we employ a trie data structure for efficiently storing and searching the symbolic representations, and investigate greedy search, bounding and sampling techniques for selecting features.
We also investigate the impact of the type (e.g. SAX or SFA) and the number of symbolic representations (e.g., multiple SFA representations by changing parameters) used for selecting features.
For the third stage, we employ an efficient linear classifier based on logistic regression. While the choice of the learning algorithm does not depend on the previous two stages, we select logistic regression for its scalability, accuracy and the benefit of model transparency and calibrated prediction probabilities, which can benefit some follow up steps such as classifier interpretation. For example, in the MrSEQL approach \citep{LeNguyen2019}, the symbolic features selected by the  logistic regression model can be maped back to the time series to compute a saliency map explanation for the classifier prediction.
A schematic representation of the MrSQM approach is given in Figure~\ref{fig:mrsqm}.

\subsection{Symbolic Representations of Time Series}
\label{method:symbolic}
While SAX and SFA are two different techniques to transform time series data to symbolic representations, both can be summarized in three steps:

\begin{itemize}
\item Use a sliding window to extract segments of time series (parameter $l$: window size).
\item Approximate each segment with a vector of smaller or equal length (parameter $w$: word size).
\item Discretise the approximation to obtain a symbolic word (e.g., \textit{abbacc}; parameter $\alpha$: alphabet size).
\end{itemize}

\begin{center}
\begin{figure}[ht]
\includegraphics[width=0.9\linewidth]{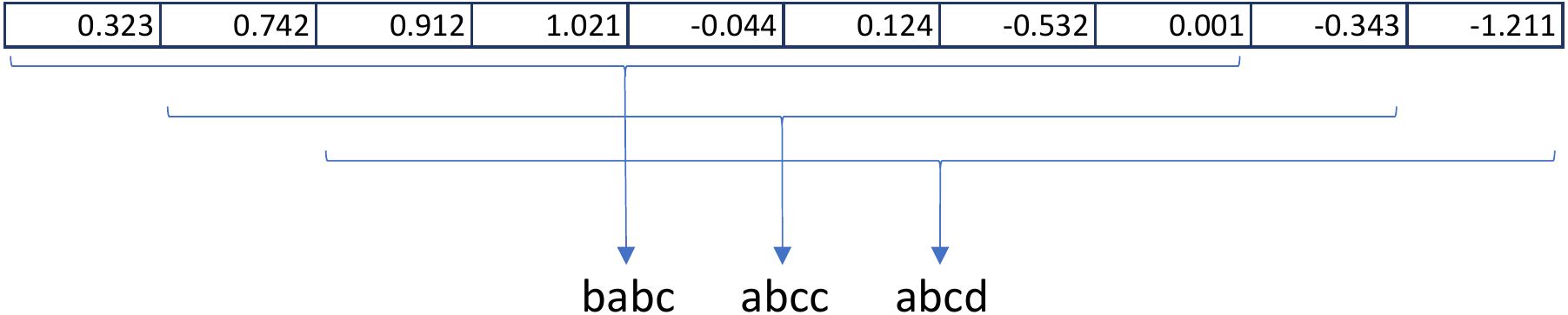}
\caption{Example symbolic transform using a sliding window over the time series.}
\label{fig:jump-ts} 
\end{figure}
\end{center}
As a result, the output of transforming a time series is a sequence of symbolic words (e.g., \textit{abbacc aaccdd bbacda aacbbc}). Figure \ref{fig:jump-ts} shows an example symbolic transform applied to a time series, and the resulting sequence of symbolic words.

The main differences between SAX and SFA are the approximation and discretisation techniques, which are summarized in Table~\ref{tbl:sax_vs_sfa}. The SAX transform works directly on the raw numeric time series, in the time domain, using an approximation called iecewise Aggregate Approximation (PAA). The SFA transform builds on the Discrete Fourier Transform (DFT), and then discretisation in the frequency domain. Hence these two symbolic transforms should capture different types of information about the time series structure.
Each transform results in a different symbolic representation, for a fixed set of parameters $(l, w, \alpha)$. This means that for a given type of symbolic transform (e.g., SAX), we can obtain multiple symbolic representations by varying these parameters. This helps in capturing the time series structure at different granularity, e.g., by varying the window size $l$ the symbolic words capture more detailed or higher level information about the time series. 

\begin{table}[ht]
\centering
\caption{Differences between SAX and SFA symbolic transforms. $N$ is the number of time series and $L$ is the length of time series. Piecewise Aggregate Approximation (PAA) is used in SAX and Discrete Fourier Transform (DFT) is used in SFA.}
 \begin{tabular}{|l| c c |} 
 \hline
 & SAX & SFA \\
 \hline
Approximation & PAA & DFT \\
Discretisation & equi-probability bins  & equi-depth bins  \\
Complexity & $\mathcal{O}(NL)$ & $\mathcal{O}(NL\log{}L)$ \\
 \hline
 \end{tabular}
 \label{tbl:sax_vs_sfa}
\end{table}

 MrSQM generates $k \times log(L)$ representations by randomly sampling values for $(l,w,\alpha)$ from a range of values, as shown in Table \ref{tbl:par_exp}. Parameter $k$ is a controlling parameter that can be set by the user. 

\begin{table}[h!]
\centering
\caption{An example of parameter sampling for a dataset of time series length $L = 64$.}
 \begin{tabular}{| l |c | c |} 
 \hline
 & MrSQM  & MrSEQL \\
 \hline
window size & $2^{3 + i/k}$ for $i$ in $(0,1, \dots log(L))$ & 16,24,32,40,48,54,60,64 \\
word length & 6,8,10,12,14,16 & 16\\
alphabet size & 3,4,5,6 & 4\\

 \hline
 \end{tabular}
 \label{tbl:par_exp}
\end{table}

 In comparison, the MrSEQL classifier creates approximately $\sqrt{L}$ symbolic  representations for each time series (where $L$ is the length of the time series). It does this by fixing the values for the alphabet to $\alpha=4$ and word size $w=16$, starting from a fixed window length of size $l=16$ and increasing the length by $\sqrt{L}$ to obtain each new symbolic representation. The number of representations for MrSEQL is thus  automatically set by the length of the time series $L$. This new sampling strategy helps MrSQM to scale better for long time series. Moreover, MrSQM samples the window size using an  exponential scale, i.e., it tends to choose smaller windows more often, while MrSEQL gives equal importance to windows of all sizes (see Table \ref{tbl:par_exp} for an example).


\subsection{Feature Selection Methods for Symbolic Transformations of Time Series }

Once we have the time series transformed to sequences of symbolic words, we can investigate different methods for feature selection. The focus here is on efficient methods that can exploit the sequence structure and feature quality bounding techniques for fast navigation of the feature space. 

\subsubsection{Supervised Feature Selection}
\begin{figure}[ht]
\centering
\includegraphics[width=0.6\linewidth]{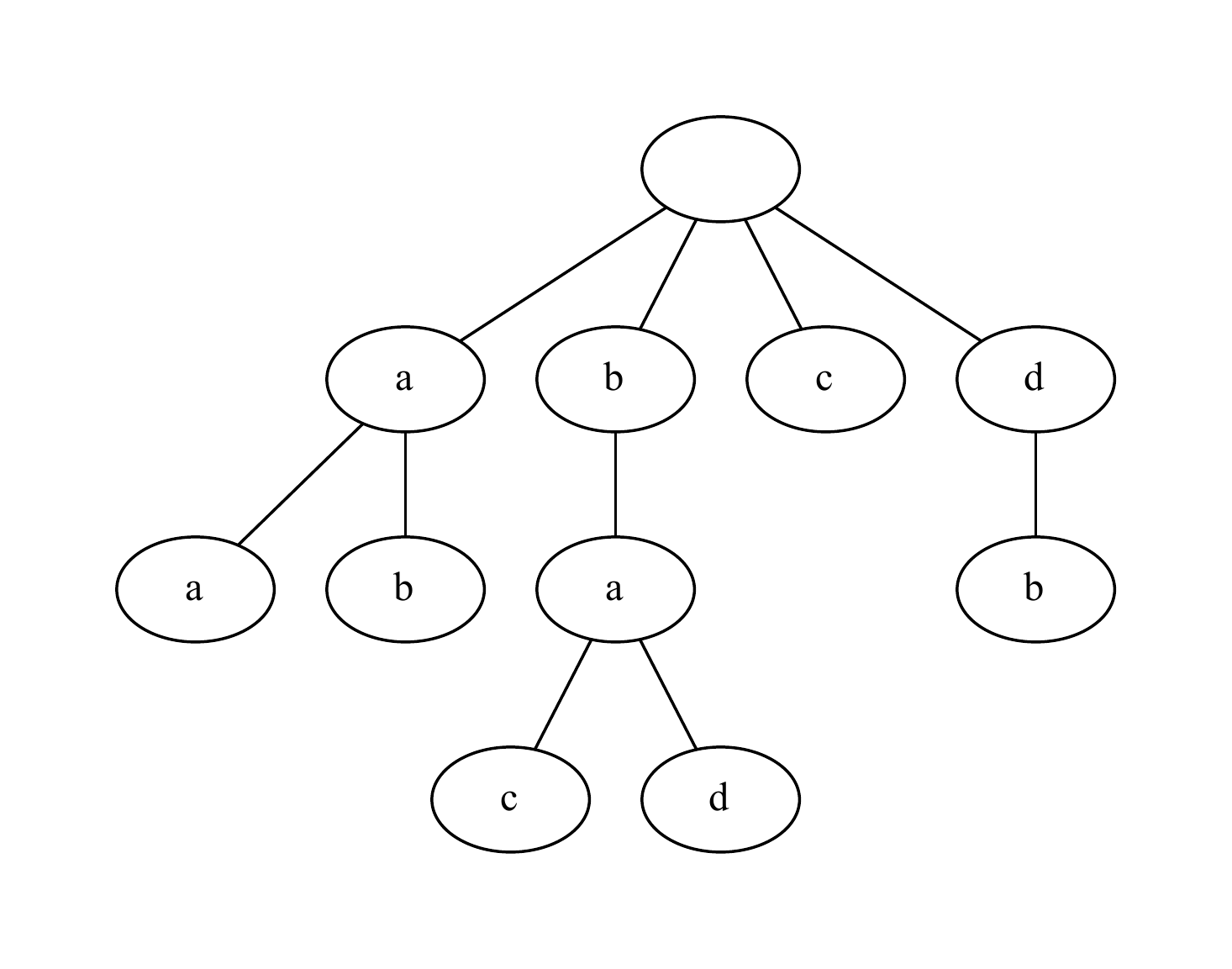}
\caption{An example of a trie data structure for fast symbolic feature search.}
\label{fig:trie_exp} 
\end{figure}

A well-known method to rank a set of feature candidates is the Chi-square test (Chi2). The method computes a statistic for each feature given by:
\begin{equation}
\chi^2 (O_1,O_2,\dots,O_n) = \sum_{k=1}^{n} \frac{(O_k - E_k)^2}{E_k}
\end{equation}
where $O_k$ is the observed frequency and $E_k$ is the expected frequency of a feature in class $c_k$. If the observed and expected frequency are similar, then the critical value approaches zero which suggests higher independence between the feature and the class. In our approach, we consider each sub-word found in the symbolic representation, as a candidate feature. It is thus very expensive to exhaustively evaluate and rank all the candidates using the Chi2 score. Fortunately, the Chi2 statistic has an upper bound~\citep{siegfried2006} which is particularly useful for  sequence data. The bound is given by:
\begin{equation}
 \label{eq:chi2_bound}
\chi^2 (O^\prime_1,O^\prime_2,\dots,O^\prime_n) \le \max ( \chi^2 (O_1,0,\dots,0), \\ \dots,   \chi^2 (0,0,\dots,O_n) )
\end{equation}
where $O^\prime_k \le O_k$ for $k = 1,2,\dots,n$. 
The anti-monotonicity property of sequence data guarantees that a sequence can only be as frequent as its prefix. As a result, the Chi2 score of a candidate feature can be bounded early by examining its prefix using Equation~\ref{eq:chi2_bound}.
To explore the all-subsequence space efficiently and utilize the bound effectively, we use a trie to store subsequences.  
A trie is a tree data structure where an edge corresponds to a symbol and a node to a subsequence constructed by concatenating all the edges on the path from root to that node. As a result, a parent's subsequence is always the prefix of its child's subsequence. Figure~\ref{fig:trie_exp} shows an example trie data structure used for navigating the feature space. 

Each trie node stores the inverted index (list of locations) of the corresponding subsequence. 
In this way, the child nodes can be created quickly by iterating through the list of locations.
For the sake of simplification, assume we are searching for discriminative subsequences on a two-class sequence data set. At node $i$ of the trie, MrSQM finds the sequence $s_i$. The frequency of $s_i$ is $O^i_1$ in class 1 and $O^i_2$ in class 2. Therefore, the Chi2 critical value of $s_i$ is $\chi^2 (O^i_1,O^i_2)$. For every descendant $j$ of node $i$, its sequence $s_j$ is a subsequence of $s_i$, hence its frequencies $O^j_1$ and 
$O^j_2$ are lesser or equal to $O^i_1$ and $O^i_2$ respectively. This satisfies the condition in Equation~\ref{eq:chi2_bound}, which means:

\begin{equation} \label{eq:chi2_bound_exp}
\chi^2 (O^j_1,O^j_2) \le \max ( \chi^2 (O^i_1,0), \chi^2 (0,O^i_2))
\end{equation}

The right hand side of the inequality is the bound for all descendants of node $i$. If the bound is less than a threshold, then none of the future descendants has the potential to be selected. In this case, they can safely be pruned from the trie. The threshold is set based on the number of features to be selected. This example can be easily generalised to multi-class data. The pseudo-code for this greedy feature selection method is given in Algorithm~\ref{alg:bound}.


\begin{algorithm}[ht]
\caption{Fast selection of top $k$ subsequences from symbolic sequences, based on a multi-class Chi2 score bound and branch-and-bound approach.}
\begin{algorithmic}[1]
\State Input: Sequence data S, threshold = 0
\State F = [ ] , unvisited = [ ],
\State Initialize trie root node
\ForAll{i in range(0,length(S))}
\State root.children[S[i]].locations.add(i)
\EndFor
\State unvisited.add(root.children)
\While{unvisited is not empty}
\State node = unvisited.pop()
\If{$Chi2(node) \geq threshold$}
\State F.add(node)
\State Sort F according to Chi2 score
\If{F.size $\geq$ k}
\State threshold = Chi2(F[k])
\EndIf
\EndIf
\If{$Chi2Bound(node) \geq threshold$}
\ForAll{loc in node.locations}
\If{loc $<$ length(S) - 1}
\State node.children[S[i+1]].locations.add(i+1)
\EndIf
\EndFor
\State unvisited.add(node.children)
\EndIf
\EndWhile
\end{algorithmic}
\label{alg:bound}
\end{algorithm}



Although there are several studies of Chi2 score bounding techniques in the pattern mining community for discriminative pattern mining (e.g., \citep{Fradkin2015,siegfried2006}, this bound was never implemented on time series data. We were also unable to find any off-the-shelf implementation for sequence data. Our adaptation of this bounding technique for time series enables us to have an efficient algorithm for symbolic time series classification. Furthermore, we adapt this bounding technique to work with both SAX and SFA sequences, thus extracting the benefit of both time domain and frequency domain symbolic features.

We note that Information Gain (IG) is an alternative measure for selecting discriminative features. IG describes how well a predictor can split the data along the target variable. It also has a similar bound which is applicable in sequence data. However, the bound only works with 2-class problems \citep{Fradkin2015}. In our experiments we found no significant difference in using IG versus the Chi2 score.
The multiclass Chi2 bound allows us to find the best discriminative features very fast and guarantees that the selected features are optimal under this feature selection method (i.e., there are no other features with higher Chi2 score).

\subsubsection{Unsupervised Feature Selection}

Sequence data usually consists of a high amount of dependent subsequences (e.g., $aaab$ always occurs wherever $aaabb$ occurs). If one of them is discriminative according to the Chi2 score, the other will also be discriminative and they will likely be selected together.
The Chi2 test or any similar supervised method which ranks features independently, tends to be vulnerable to this collinearity problem. In other words, the top ranking features can be highly correlated and, as a result, induce \textit{redundancy} in the set of selected features. Random candidate selection can increase \textit{diversity} in the feature set.
This method simply selects features from a candidate set in a random fashion. 
Because this method is inexpensive, it can be applied on the set of all subsequences found in the sequence data. For the symbolic representation of time series, each symbolic sub-word is a candidate feature. The random feature sampler simply samples the index of a time series, the location within the time series and the sub-word length. In our experiments we find that with a large enough number of features sampled from multiple symbolic representations, and a linear classifier, we can also achieve high accuracy with this simple method. 


\subsubsection{Hybrid Feature Selection}

Because both supervised feature selection (which finds useful, but redundant features) and unsupervised selection (finds noisy, but diverse features) have advantages and disadvantages, they can complement each other in a hybrid approach. In this approach, the output feature set of one method is used as the candidate set of the other method. MrSQM implements two hybrid approaches: (1) supervised feature selection using the Chi2 score, then unsupervised filtering through random sampling, (2) unsupervised feature selection through random sampling, then supervised filtering using the Chi2 score. For all the feature selection methods, it is important that the symbolic transform of the time series captures useful information about the time series structure in the symbolic sequence. We discuss in Section \ref{sec:eval} the importance of parameter sampling for the symbolic transform, as well as the number and diversity of features selected from multiple symbolic transforms of the same time series.

\subsection{MrSQM Classifier Variants}

We implement these approaches for feature selection in MrSQM, resulting in four classifier variants. Note that in all the methods, all the symbolic sub-words in a symbolic representation are considered as candidate features. The features extracted from each symbolic representation are then concatenated in a single feature space which is used to train a linear classifier (logistic regression).

\begin{itemize}
    \item MrSQM-R: Unsupervised feature selection by random sampling of features (i.e., subwords of symbolic words, either SAX or SFA). 
    \item MrSQM-RS: Hybrid method, the unsupervised MrSQM-R produces candidates and a follow-up supervised Chi2 test filters features based on the Chi2 score.
    \item MrSQM-S: Supervised feature selection by pruning the all-subsequence feature space with the Chi2 bound presented in Section \ref{section:method} and selecting the optimal set of top $k$ subsequences under the Chi2 test.
    \item MrSQM-SR: Hybrid method, the supervised MrSQM-S produces candidate features ranked by Chi2 score, then a random sampling step filters those features. 
\end{itemize}

\textbf{Time Complexity.} All our classifier variants have a time complexity  
dominated by the symbolic transform time complexity. In our case, the 
SFA transform, which is $\mathcal{O}(NL\log{}L)$ is the dominant factor. This is repeated $\mathcal{O}(\log{}L)$ times (the number of symbolic representations being generated) hence the overall time complexity of MrSQM is $\mathcal{O}(NL(\log{}L)^2)$. Although SFA has a time complexity of  $\mathcal{O}(NL\log{}L)$, we build our SFA implementation on the latest advances for efficiently computing the Discrete Fourier Transform\footnote{FFTW is an open source C library for efficiently computing the Discrete Fourier Transform (DFT): \url{https://www.fftw.org}}, which results in significant time savings as compared to older SFA implementations.

\section{Evaluation}
\label{sec:eval}

\subsection{Experiment Setup}

We ran experiments on 112 fixed-length univariate time series classification datasets from the new UEA/UCR TSC Archive. MrSQM also works with variable-length time series, without any additional steps being required (i.e., once it is supported by the input file format). Since the majority of state-of-the-art TSC implementations only support fixed-length time series, for comparison, we have also restricted our experiments to fixed-length datasets.

MrSQM is implemented in C\texttt{++} and wrapped with Cython for easier usability through Python. 
For our experiments we use a Linux workstation with an Intel Core i7-7700 Processor and 32GB memory. 
To support the reproducibility of results, we have a Github  repository\footnote{\url{https://github.com/mlgig/mrsqm}} 
with all the code and results.  
 All the datasets used for experiments are available from the UEA/UCR TSC Archive website\footnote{\url{https://http://timeseriesclassification.com}}. We also obtained the accuracy results for some of the existing classifiers from the same website. For the classifiers that we ran ourselves, we have used the implementation provided in the sktime library\footnote{\url{https://www.sktime.org/en/stable/get_started.html}}.

For accuracy comparison of multiple classifiers, we follow the recommendation in \citep{Demsar:2006,garcia-extension,JMLR:v17:benavoli16a}. The accuracy gain is evaluated using a Wilcoxon signed-rank test with Holm correction and visualised with the critical difference (CD) diagram. The CD shows the ranking of multiple methods with respect to their average accuracy rank computed across multiple datasets. 
Methods that do not have a  statistically significant difference in rank, are connected with a thick horizontal line. 
For computing the CD we use the R library \textit{scmamp}\footnote{\url{https://github.com/b0rxa/scmamp}} \citep{scmamp}. While CDs are a very useful visualization tool, they do not tell the full story since minor differences in accuracy can lead to different ranks. In order to get a more complete view of results, we supplement the CD with tables and pairwise scatter plots for a closer look at the accuracy and runtime performance.

    

\subsection{Sensitivity Analysis}

\subsubsection{Comparing MrSQM Variants}
\begin{figure}[ht]
\centering
\includegraphics[width=\linewidth]{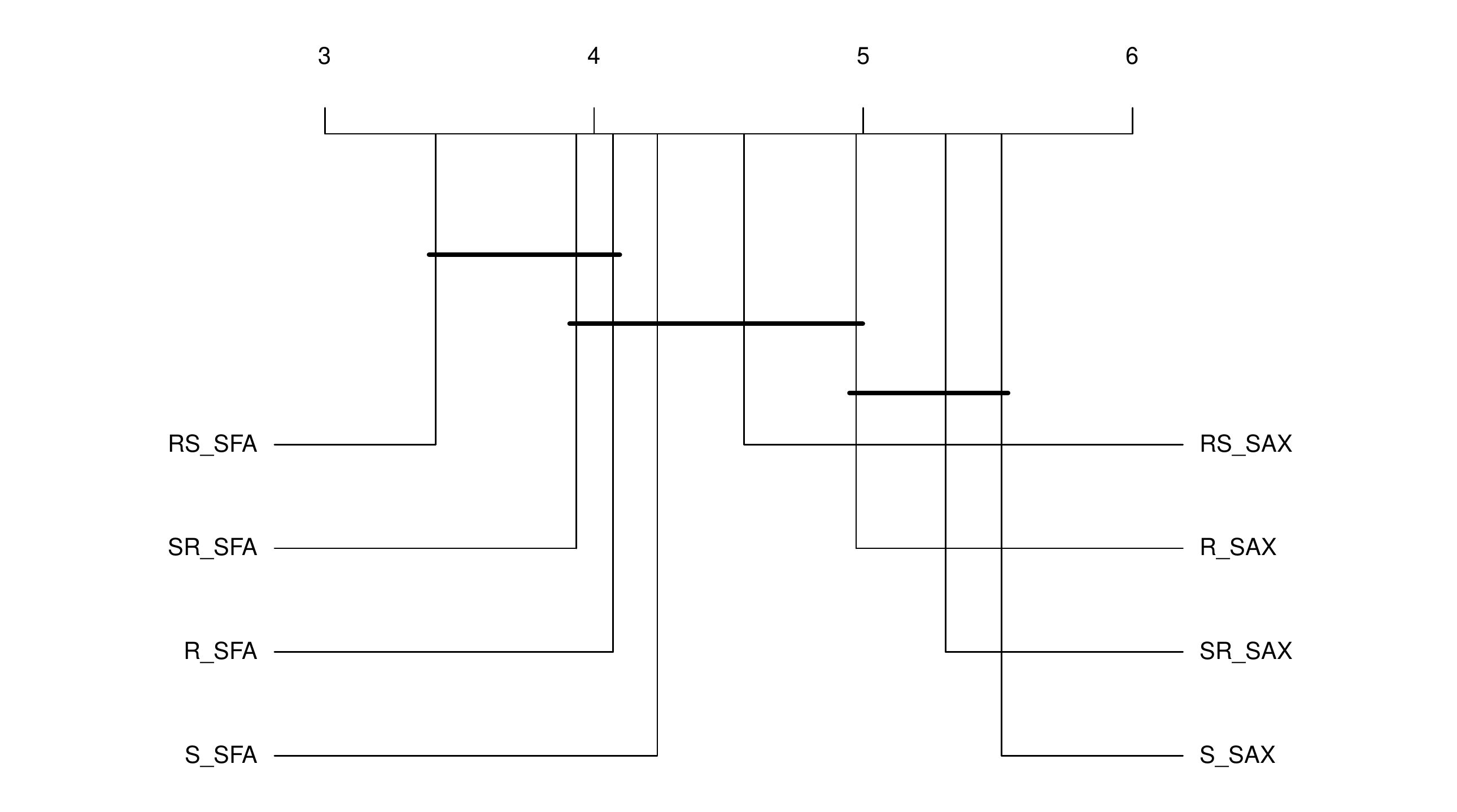}
\caption{Comparison between eight variants of MrSQM for different symbolic representations and feature selection strategies.}
\label{fig:fs_vs_st} 
\end{figure}

In this section, we investigate the two main components of MrSQM: symbolic transformation and feature selection. For symbolic transformation we consider either the SAX or the SFA transform. Feature selection considers the four strategies described before: R, S , SR, and RS. In Figure~\ref{fig:fs_vs_st} we compare the eight different combinations, i.e., for each type of transform, we evaluate how the feature selection methods behave. Note that for this experiment, we set $k=1$ as described in Section \ref{method:symbolic}. The type of feature selection is denoted before the transform, i.e., $RS\_SAX$ denotes the MrSQM variant that uses SAX representations and the RS strategy (random feature selection, followed by filtering with a supervised Chi2 test). For each transform, 500 features are selected. The number of features per representation does not seem to play a major role, once we have a few hundred features. In our experiments we varied the number of features from 100 to 2000, and from 500 onward the accuracy does not change significantly, hence the default is set to 500.

\begin{table}[h!]
\centering
\caption{Time (minutes) comparison for SAX versus SFA combined with different feature selection strategies with $k=1$.}
\begin{tabular}{l|llll}
\hline
Symbolic Transform  & \multicolumn{4}{c}{Feature Selection}\\
                    & R     & RS    & SR      & S \\\hline
SAX               & 27    & 28    & 28      & 27  \\    
SFA               & 17    & 22    & 19        & 20  \\
\hline
\end{tabular}
\end{table}

It is clear from this experiment (Figure \ref{fig:fs_vs_st}) that the SFA symbolic transform is generally superior to SAX, for all feature selection variants. On the other hand, among the feature selection methods, the RS strategy seems to be more effective than the other three. All of these variants are very fast, totaling around 20 minutes for training and predicting on the entire 112 datasets using SFA, across all feature selection methods. For SAX, all four methods take about 30 minutes to complete training and prediction on 112 datasets.
In \citep{LeNguyen2019} it was shown that for the MrSEQL classifier expanding the feature space by adding more symbolic representations can improve the accuracy. In the next experiment, we investigate this hypothesis. Generally, there are two ways to add more representations: adding representations of the same type or adding representations of a different type.

Since the SFA and RS variants are more accurate than the others, from this point onward they will be our default choices for the experiments unless stated otherwise.

\subsubsection{Comparing MrSQM to MrSEQL}
\label{subsec:mrsqmvsmrseql}

\begin{figure}[ht]
\centering
\includegraphics[width=0.9\linewidth]{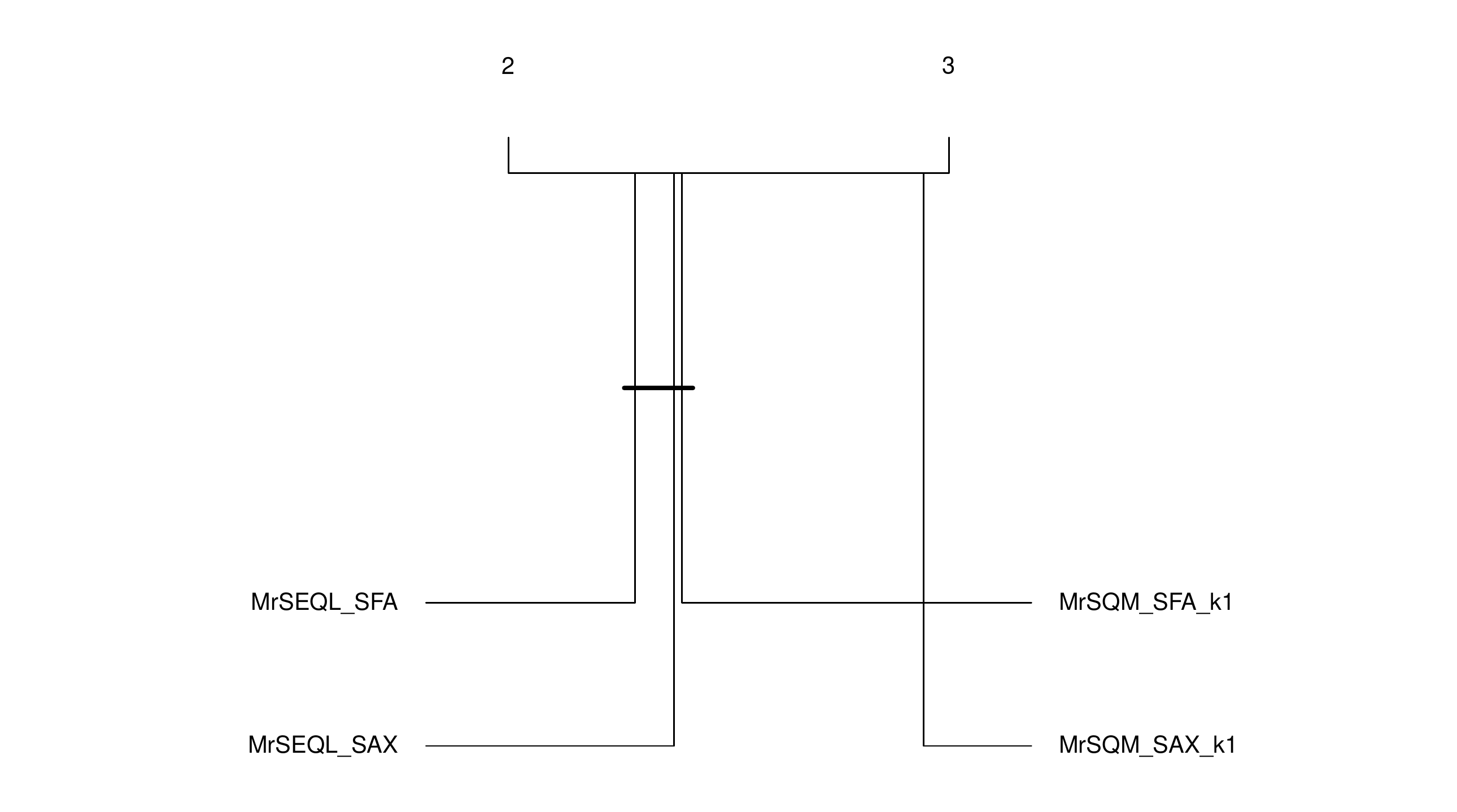}
\caption{Comparison between variants of MrSQM and MrSEQL classifiers.}
\label{fig:sqm_vs_seql} 
\end{figure}

Figure \ref{fig:sqm_vs_seql} shows the comparison between MrSEQL and MrSQM with either SAX or SFA representations. The SAX variant of MrSQM appears to be significantly less accurate than the other three, while the SFA variant of MrSQM is similarly accurate to the MrSEQL variants. However, we note that these variants of MrSQM took less than 30 minutes to complete training and prediction on the entire benchmark (20 minutes for SFA and 30 minutes for SAX), while MrSEQL took more than 3 hours for SFA and about 7 hours for SAX (see Figure \ref{fig:sqm_vs_seql_scatter} for details on runtime). This significant speedup is partly due to the change in sampling symbolic parameter values, coupled with the more significant change in the feature selection strategy implemented in MrSQM which is more efficient than the feature selection implemented in MrSEQL.  

\subsubsection{Parameter Sampling for the Symbolic Transform}

In this set of experiments, we study the impact of the symbolic transformation in terms of both quality and quantity of representations.
\begin{figure}[ht]
\centering
\includegraphics[width=\linewidth]{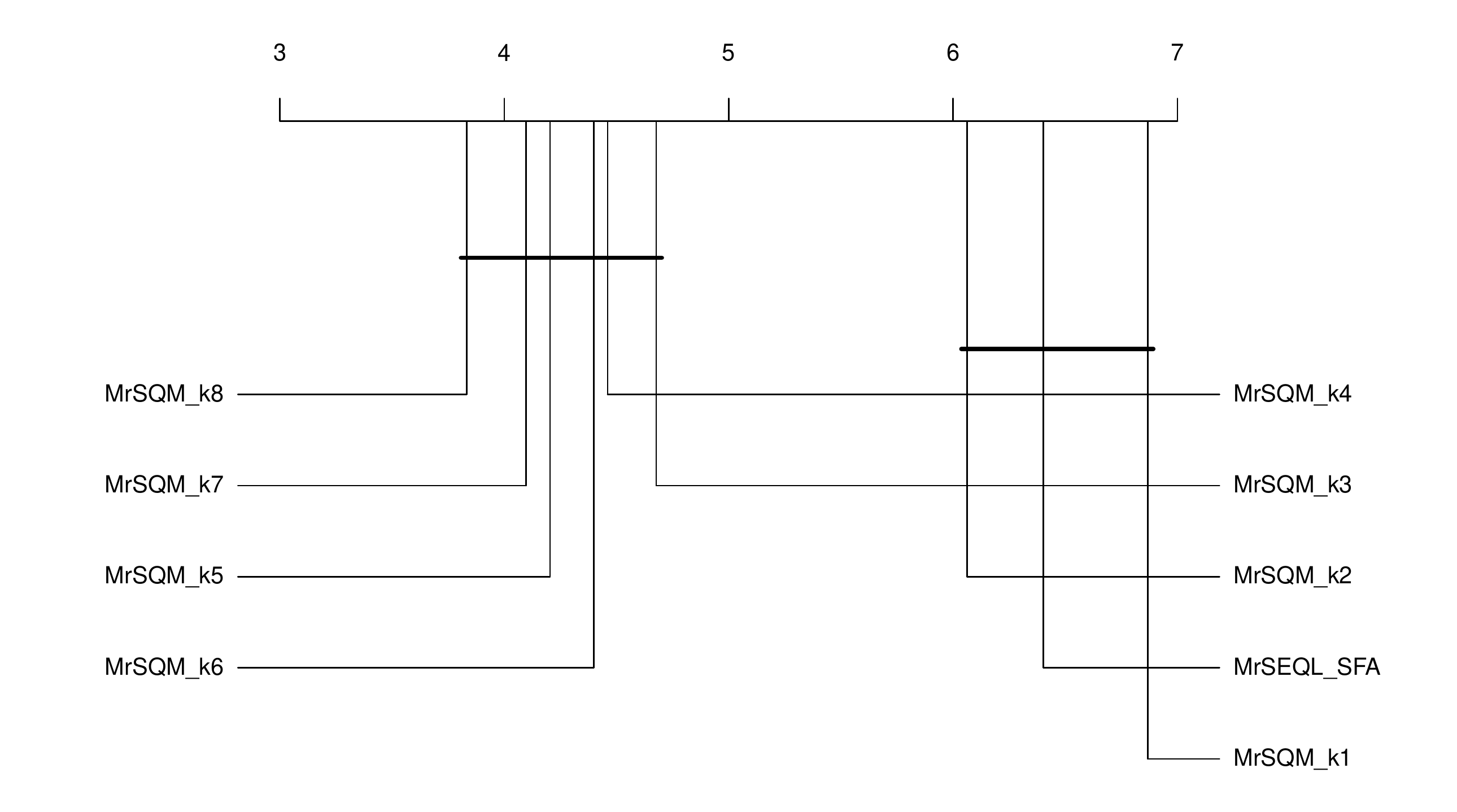}
\caption{Comparison of average accuracy rank for MrSQM-SFA variants at variable $k$ and MrSEQL-SFA as baseline.}
\label{fig:params} 
\end{figure}
Figure \ref{fig:params} shows results comparing different numbers of SFA representations (with $k$ varying from 1 to 8), when using the RS feature selection strategy for MrSQM. It also includes a comparison to the MrSEQL classifier restricted to only using SFA features, in order to directly compare the accuracy and speed, using the same type of representation.
The results show that adding more symbolic representations by varying the control parameter $k$ can benefit MrSQM, albeit with the cost of extra computation reflected in the runtime. In addition, MrSQM at $k=3$ is already significantly more accurate than MrSEQL, while still being faster. 

\begin{figure}[ht]
\centering
\includegraphics[width=\linewidth]{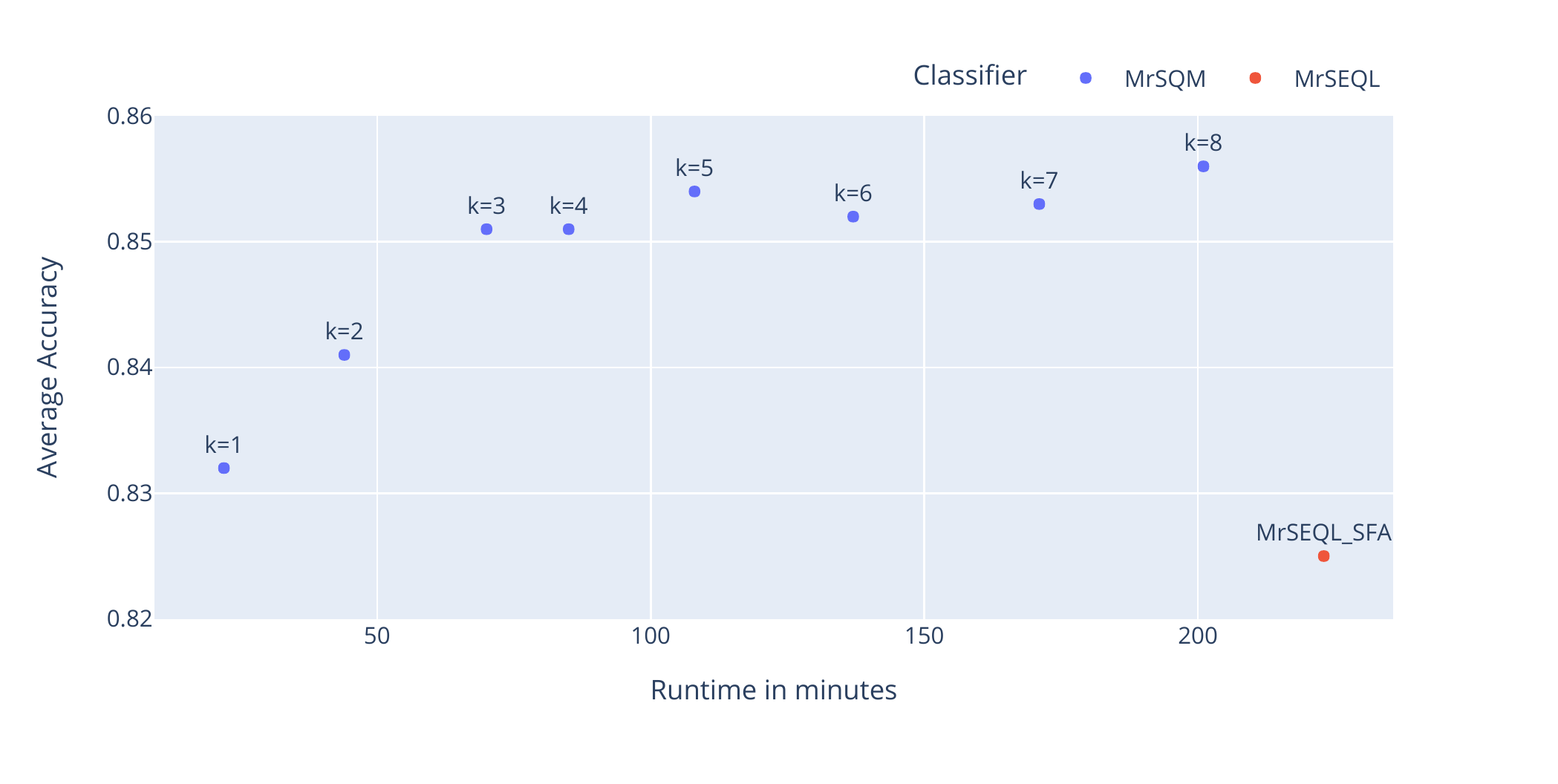}
\caption{Comparison of average accuracy and total training and prediction time (minutes) for MrSQM-SFA variants at varying $k$ and MrSEQL variants as baseline.}
\label{fig:sqm_vs_seql_scatter} 
\end{figure}

In Figure \ref{fig:sqm_vs_seql_scatter} we present a comparison of the accuracy and runtime of MrSQM (for different values for $k$) and the MrSEQL classifier.
Overall, the MrSQM variant with $k=5$ seems to achieve a good trade-off between accuracy and speed, taking slightly over 100 minutes total time.

\subsubsection{Hybrid MrSQM: Combining SAX and SFA Features}

\begin{figure}[ht]
\centering
\includegraphics[width=\linewidth]{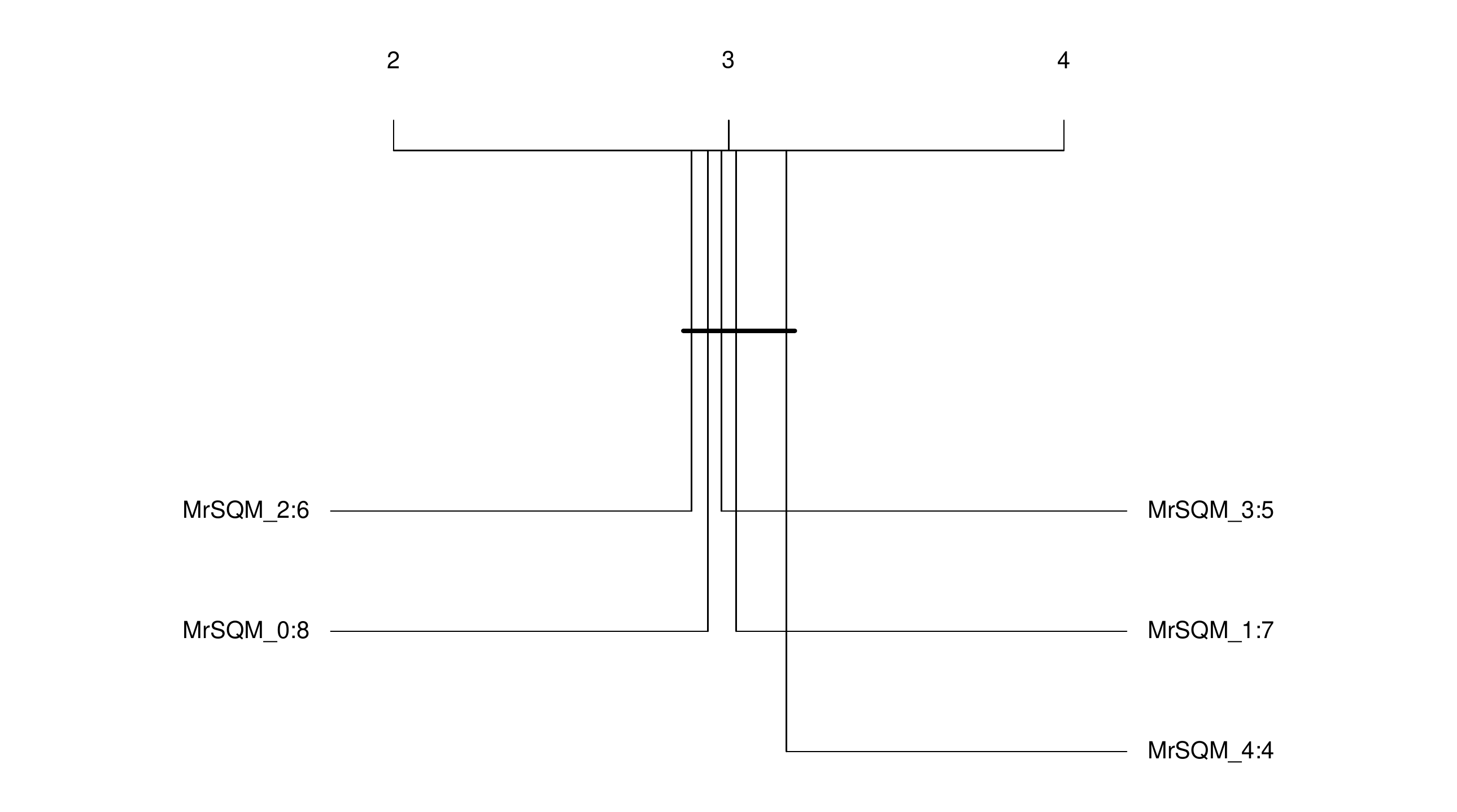}
\caption{Comparison between variants of MrSQM with different ratios of SAX and SFA representations. $k_1:k_2$ means MrSQM generates $k_1 \times log(L)$ SAX representations and $k_2 \times log(L)$ SFA representations.}
\label{fig:hybrid} 
\end{figure}

\begin{figure}[ht]
\centering
\includegraphics[width=0.95\linewidth]{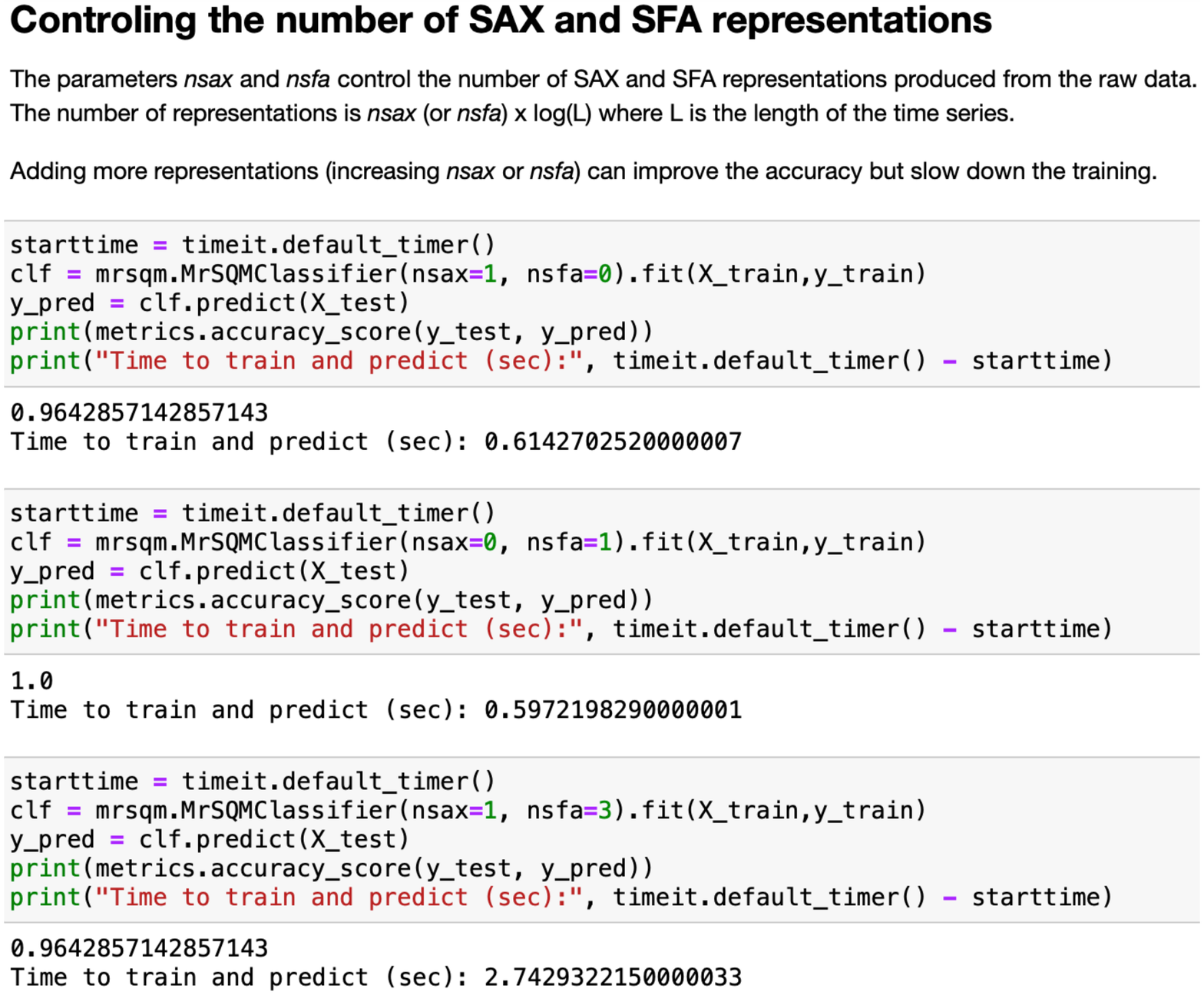}
\caption{Example Python code for setting the type and number of symbolic representations for the MrSQM classifier on the Coffee dataset.}
\label{fig:mrsqm-example-run} 
\end{figure}

In this experiment we explore the option of combining SAX and SFA feature spaces. \cite{LeNguyen2019} found that the combination of SAX and SFA features (with a 1:1 ratio) is very effective for the MrSEQL classifier. For MrSQM, we do not find the same behaviour when combining the two types of representations.  Figure~\ref{fig:hybrid} shows that the MrSQM variant that only uses the SFA transform is as effective as when using combinations of SAX and SFA representations in different ratios.
These results suggest that, to maximise accuracy and speed, the recommended choice of symbolic transformation for MrSQM is SFA. However, it is worth noting that in practice this choice can depend on the requirements of the application. Across the 112 datasets that come from a wide variety of domains, SFA seems to be outperforming the SAX transform in both accuracy and speed. Nevertheless, for many datasets, SAX and SFA models have similar accuracy. Hence, it makes sense to enable the user to select the type and the number of symbolic transforms to be used in their application.
Furthermore, like MrSEQL, the MrSQM classifier can produce a saliency map for each time series, from models trained with SAX features. This can be valuable in some scenarios where classifier interpretability is desirable, and MrSQM enables the user to select the transform that best fits their application scenario. In Figure \ref{fig:mrsqm-example-run} we show an example for how the user can control the type and number of SAX or SFA representations for the Coffee dataset. Further examples are provided in the Jupyter notebook\footnote{\url{https://github.com/mlgig/mrsqm/blob/main/example/Time_Series_Classification_and_Explanation_with_MrSQM.ipynb}} that accompanies our open source code for MrSQM.

\subsection{MrSQM versus State-of-the-art Symbolic Time Series Classifiers}

\begin{figure}[ht]
\centering
\includegraphics[width=0.9\linewidth]{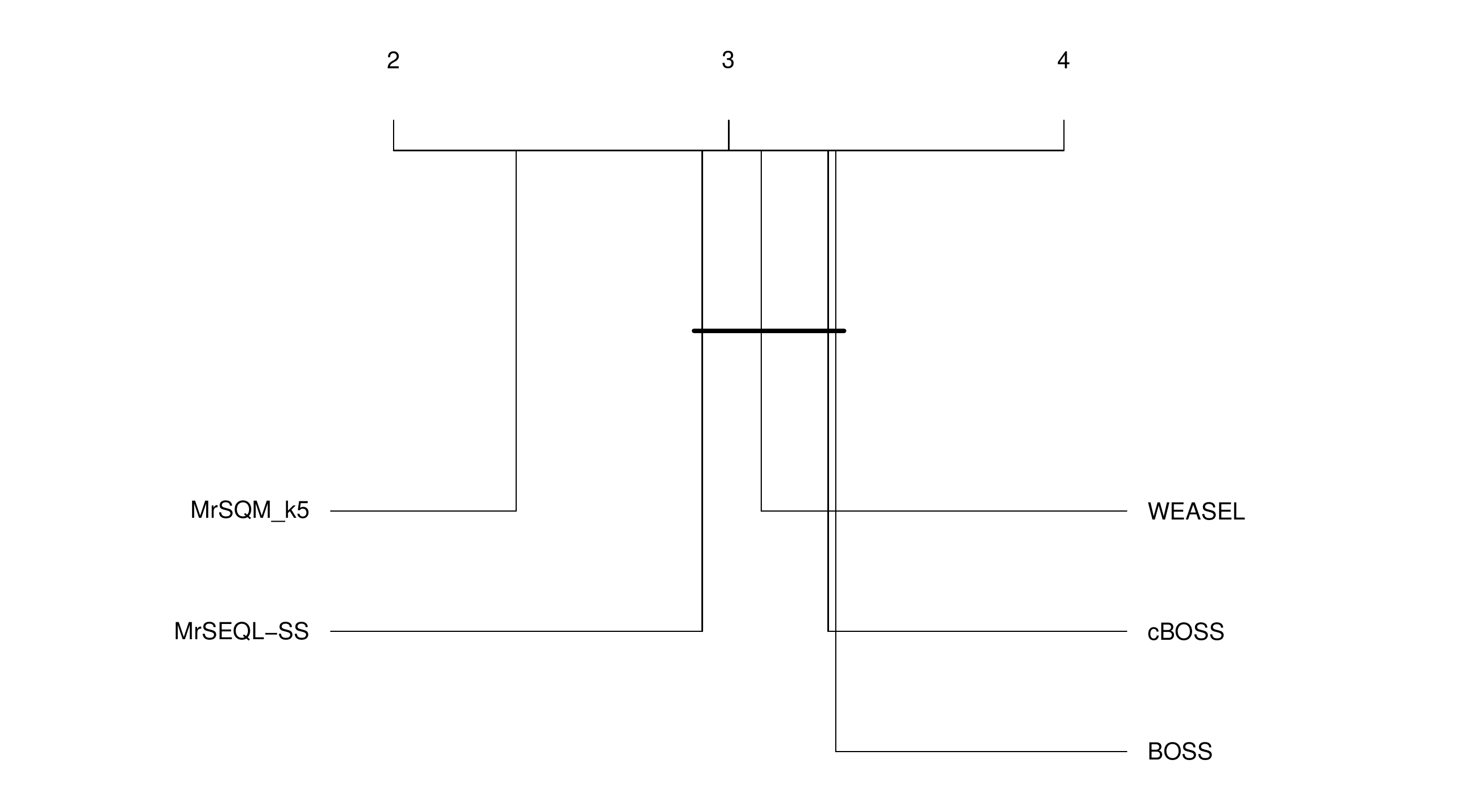}
\caption{Comparison of state-of-the-art symbolic time series classifiers: average rank order with regard to accuracy across 112 UEA/UCR TSC datasets. The leftmost method has the best average rank.}
\label{fig:symb_cd} 
\end{figure}

We compare our best classifier variant for MrSQM (MrSQM\_k5 that has a good accuracy-time trade-off as shown in Figure \ref{fig:sqm_vs_seql_scatter}) with state-of-the-art symbolic time series classifiers. This group includes WEASEL, MrSEQL, BOSS and cBOSS~\citep{rboss2019}. All five classifiers use SFA representations to extract features, while MrSEQL uses both SAX and SFA representations (Figure~\ref{fig:symb_cd}).

MrSQM has the highest average rank and is  significantly more accurate than the other symbolic classifiers. Furthermore, all the other methods require at least 5-12 hours to train, as shown in Figure \ref{fig:time_vs_acc} and results reported in  \citep{Bagnall2020ATO,LeNguyen2019}. We note that the ensemble methods (e.g., BOSS, cBOSS) are outperformed by the linear classifiers. 
With regard to runtime, as shown in Table \ref{tbl:time_sota}, MrSQM is  significantly faster than the other symbolic classifiers (MrSQM takes 1.5h to complete training and prediction, versus 10h for MrSEQL-SS).


\subsection{MrSQM versus other State-of-the-Art Time Series Classifiers}

The group of the most accurate time series classifiers that have been published to date include HIVE-COTE, TS-CHIEF, ROCKET (and its extension MiniROCKET), and InceptionTime. With the exception of ROCKET and MiniROCKET, these classifiers are very demanding in terms of computing resources. Running them on 112 UEA/UCR TSC datasets takes days and even weeks to complete training and prediction  \citep{Bagnall2020ATO,DBLP:journals/ml/MiddlehurstLFLB21}.
\begin{figure}[ht]
\centering
\includegraphics[width=\linewidth]{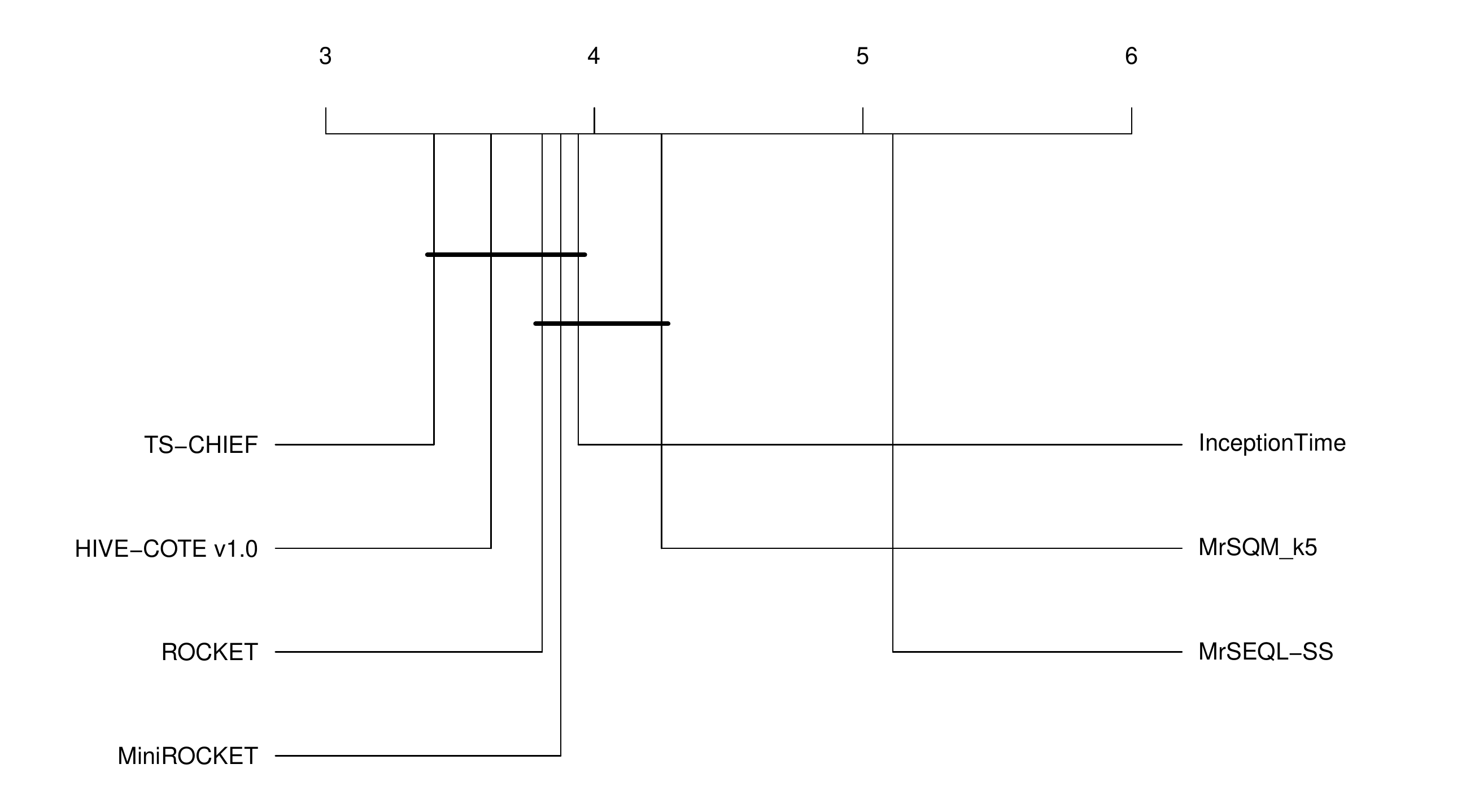}
\caption{Comparison between state-of-the-art time series classifiers and MrSQM with regard to average accuracy rank across 112 UEA/UCR TSC datasets. The leftmost method has the best average rank.}
\label{fig:sota_cd} 
\end{figure}

\begin{figure}[ht]
\centering
\includegraphics[width=\linewidth]{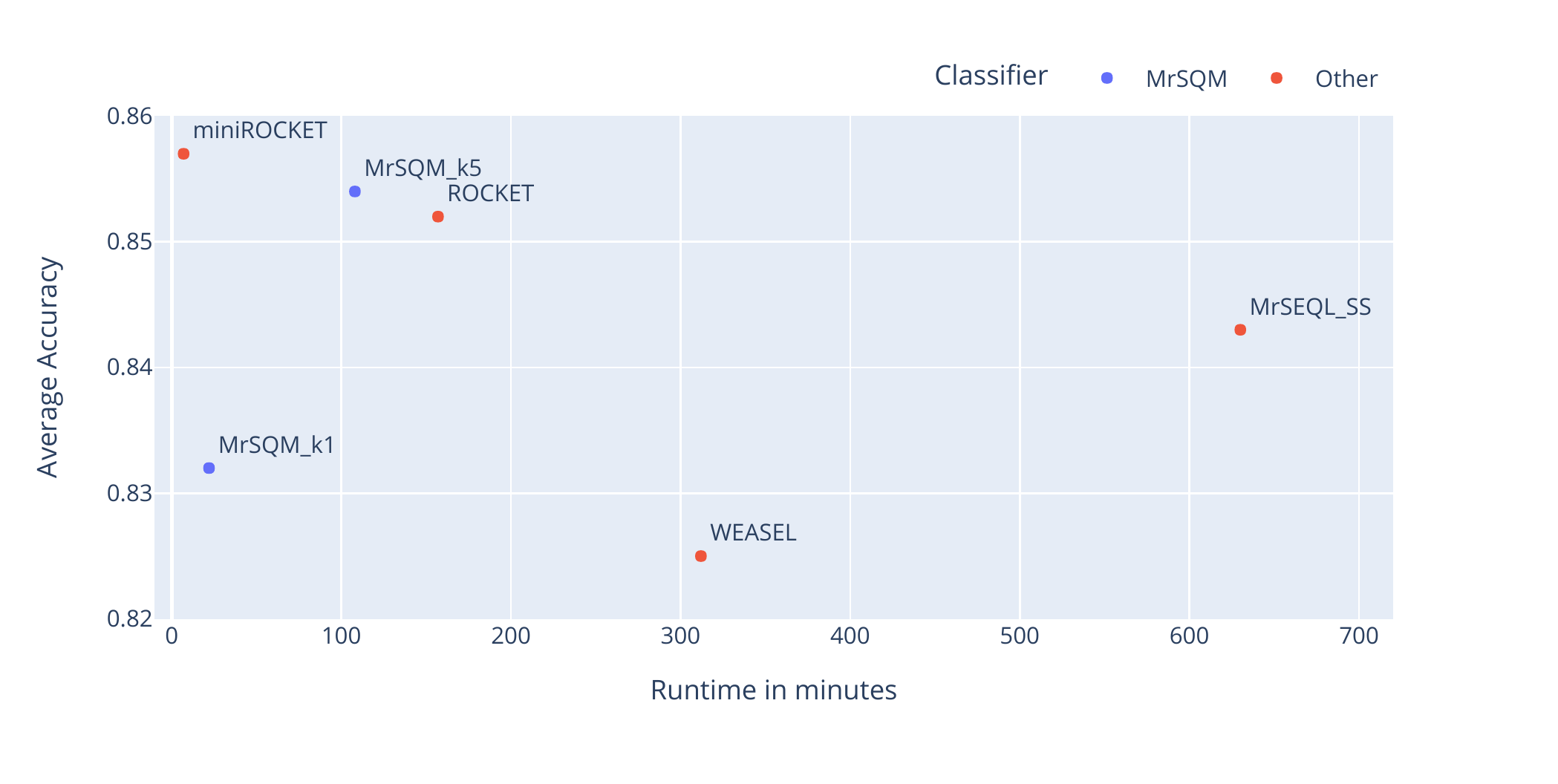}
\caption{Comparison of accuracy and total time (minutes) trade-off for MrSQM variants and state-of-the-art methods that complete training within 10 hours.}
\label{fig:time_vs_acc} 
\end{figure}

Figure \ref{fig:sota_cd} shows the accuracy rank comparison between these methods and MrSQM. Among the methods compared, only TS-CHIEF and HIVE-COTE were found to have a statistically  significant difference in accuracy. Nevertheless, these methods require more than 100 hours to complete training \citep{Bagnall2020ATO}, for a relatively small gain in average accuracy, typically of about 2\% (see Table \ref{tbl:acc_diff}). In this diagram, MrSQM is in the same accuracy group as InceptionTime, MiniROCKET and ROCKET. Neverthelss, in terms of runtime, MrSQM is in a group with ROCKET: MrSQM takes 100 minutes to complete training and prediction on 112 datasets, while ROCKET, in our run on the same machine, takes 150 minutes (see more details on runtime in Table \ref{tbl:time_sota} and Figure \ref{fig:time_vs_acc}). The MrSQM-k1 variant takes only 20 minutes  and for many datasets this variant is enough to achieve high accuracy. This variant is comparable in accuracy and runtime to the MiniROCKET classifier. In Figure \ref{fig:time_vs_acc} we show a comparison of some of these methods with regards to the accuracy versus runtime (we only include the methods that we ran ourselves on the same machine).

\begin{figure}[ht]
\centering
\includegraphics[width=\linewidth]{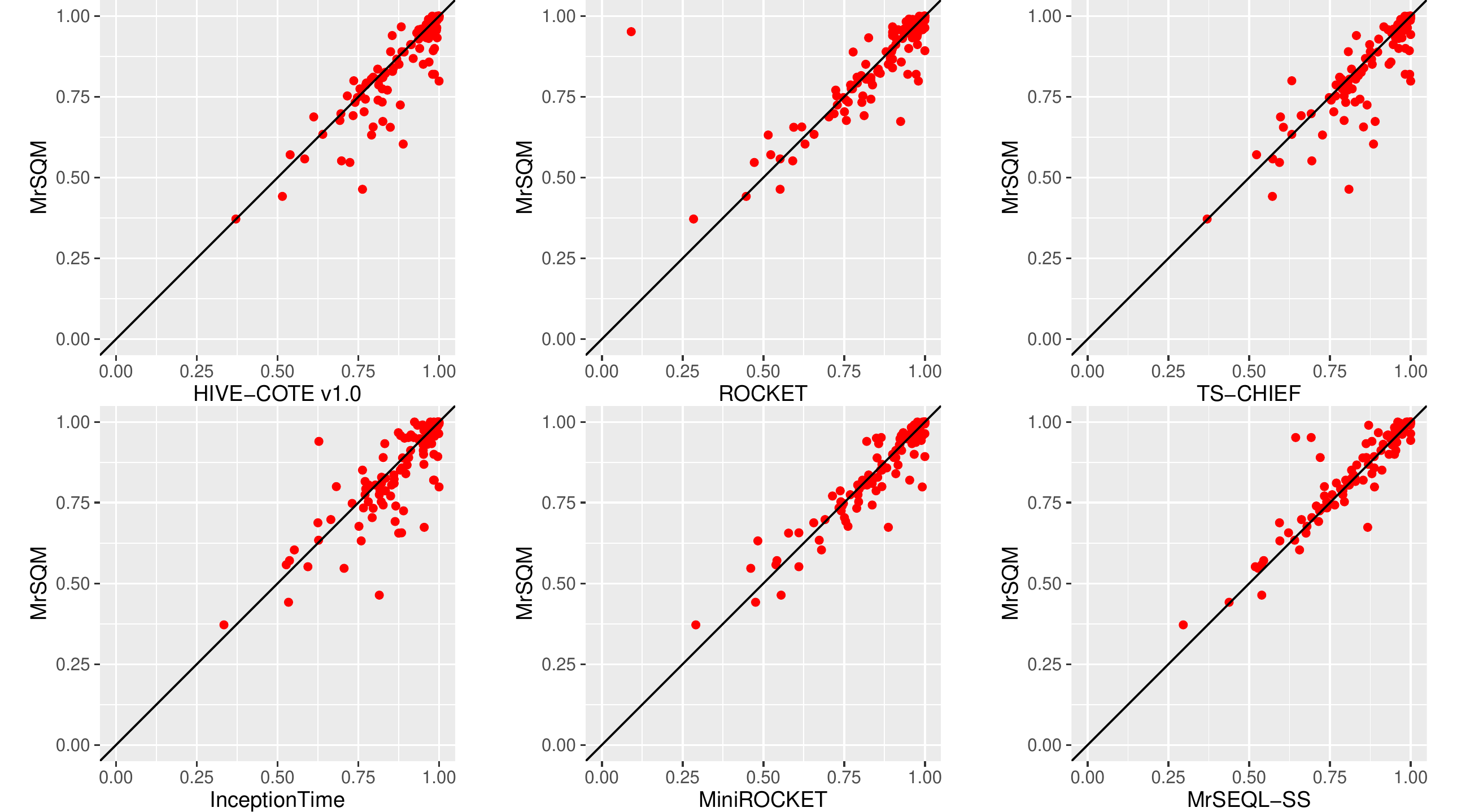}
\caption{Pairwise comparison between state-of-the-art time series classifiers and MrSQM with regard to accuracy across 112 UEA/UCR TSC datasets.}
\label{fig:sota} 
\end{figure}

Figure~\ref{fig:sota} shows the pairwise-comparison of accuracy between these methods and MrSQM. Each dot in the plot represents one dataset from the benchmark. MrSQM is more accurate above the diagonal line and highly similar methods cluster along the line. We note that the accuracy across datasets is similar for MrSQM versus ROCKET or MiniROCKET, the only other two methods in the same runtime category. 

We investigate further the difference in average accuracy for MrSQM versus the other methods. In Table~\ref{tbl:acc_diff}, we summarize the accuracy differences between MrSQM and the other classifiers. For context, in Table \ref{tbl:time_sota} we also provide the runtime for all the methods. We observe that when taken together, the average difference in accuracy and the total time to complete training and prediction over the 112 datasets, we see a clear grouping of methods. If we focus on fast methods that can complete training and prediction in a couple of hours, only the ROCKET/MiniROCKET methods and MrSQM can achieve this. If we look at the average difference in accuracy versus the other methods, there is only about 2\% difference in accuracy, for orders of magnitude faster runtime.
In the group of symbolic classifiers, MrSQM is both significantly more accurate and much faster than existing symbolic classifiers. Furthermore, while it is expected that MrSQM's results are aligned with the other symbolic methods (WEASEL, MrSEQL), it is surprising that they are also very similar to MiniROCKET (second-highest correlation) but not ROCKET (lowest correlation). Perhaps MiniROCKET is better than ROCKET at extracting frequency domain knowledge from time series data.

\begin{table}[h!]
\centering
\caption{Statistical summary of differences in accuracy between MrSQM and state-of-the-art time series classifiers.}
 \begin{tabular}{|l| c c c |} 
 \hline
 Classifiers & Mean Diff & Std Diff & Correlation \\
 \hline\hline

HIVE COTE 1.0 & 0.028 & 0.067 & 0.882 \\
TS-CHIEF & 0.026 & 0.071 & 0.866 \\
InceptionTime & 0.021 & 0.084 & 0.816 \\
ROCKET  & 0.002 & 0.099 & 0.797 \\
MiniROCKET & 0.007 & 0.052 & 0.936 \\
WEASEL & -0.025 & 0.069 & 0.91 \\
MrSEQL-SS & -0.011 & 0.055 & 0.927 \\
MrSQM\_K1 & -0.021 & 0.038 & 0.97  \\
 \hline
 \end{tabular}
 \label{tbl:acc_diff}
\end{table}

\begin{table}[h!]
\centering
\caption{Runtime of state-of-the-art classifiers for completing training and prediction over 112 datasets. For HIVE-COTE1.0 and TS-CHIEF the time is taken directly from  \citep{Bagnall2020ATO}.}
 \begin{tabular}{|l| c|} 
 \hline
 Classifier & Total (hours) \\
 \hline\hline
MiniROCKET  & 0.1 \\ 
MrSQM\_K1   & 0.3 \\
MrSQM\_K5   & 1.5 \\
ROCKET      & 2.5 \\
WEASEL      & 5 \\
MrSEQL-SS   & 10 \\
HIVE-COTE1.0    & 400 \\ 
TS-CHIEF        & 600 \\
 \hline
 \end{tabular}
 \label{tbl:time_sota}
\end{table}






\section{Conclusion}
\label{sec:conc}

In this paper we have presented MrSQM, a new symbolic time series classifier which works with multiple symbolic representations of time series, fast feature selection for symbolic sequences and a linear classifier. We showed that while conceptually very simple, MrSQM achieves state-of-the-art accuracy on the new UEA/UCR time series classification benchmark, and can complete training and prediction in under two hours on a regular computer.  
This compares very favorably to existing methods such as HIVE-COTE, TS-CHIEF and InceptionTime, which achieve only slightly better accuracy, but require  weeks to train on the same datasets and require advanced compute infrastructure. MrSQM is comparable to the recent classifier ROCKET, in regard of both accuracy and speed. 
This work has shown again that methods from the group of linear classifiers working in large feature spaces are very effective for the time series classification task.
For future work we intend to study methods to further reduce the computational complexity of symbolic transformations and extend MrSQM to work on multivariate time series classification.

\begin{acknowledgements}
This work was funded by Science Foundation Ireland through the Insight Centre for Data Analytics (12/RC/2289\_P2) and VistaMilk SFI Research Centre
(SFI/16/RC/3835).
\end{acknowledgements}

%
%

\bibliographystyle{spbasic}      
\bibliography{main}   




\end{document}